\documentclass[review]{elsarticle}

\usepackage{amsmath,amssymb,amsfonts}
\usepackage{algorithm}
\usepackage{algpseudocode}
\usepackage{graphicx}
\usepackage{tikz}
\usetikzlibrary{arrows.meta,positioning,shapes.geometric,backgrounds,fit,calc}
\usepackage{pgfplots}
\pgfplotsset{compat=1.18}
\usepackage{textcomp}
\usepackage{xcolor}
\usepackage{booktabs}
\usepackage{array}
\usepackage{multirow}
\usepackage[section]{placeins}  
\usepackage{listings}
\usepackage{hyperref}
\hypersetup{
  colorlinks=true,
  linkcolor=black,
  citecolor=black,
  urlcolor=blue,
}

\newif\ifanon
\anonfalse

\ifanon
  \journal{Data Science and Management}
\else
  \journal{Information Sciences}
\fi

\begin{document}

\begin{frontmatter}

\ifanon
\else
\title{ExecuGraph: A Multi-Agent, Execution-Grounded Framework for Reliable Backend Code Synthesis with Large Language Models\tnoteref{t1}}
\tnotetext[t1]{Source code, configurations, benchmark definitions, and per-trial result logs that produce every numeric claim in this paper are available at \url{https://github.com/rohithreddybc/multi-agent-dsa-backend-logic-synthesis}.}

\author[kitsaiml]{L.~Sai~Deekshith}
\author[kits]{A.~Jothi~Prabha}
\author[indep]{Rohith~Reddy~Bellibatlu}
\author[bu]{Manpreet Singh\corref{cor1}}
\ead{manni@bu.edu}

\affiliation[kitsaiml]{organization={Department of Computer Science and Engineering (AI \& ML), Kakatiya Institute of Technology and Science},
  city={Warangal}, state={Telangana}, country={India}}
\affiliation[kits]{organization={Department of Computer Science and Engineering, Kakatiya Institute of Technology and Science},
  city={Warangal}, state={Telangana}, country={India}}
\affiliation[indep]{organization={Independent Researcher}, country={India}}
\affiliation[bu]{organization={Boston University},
  city={Boston}, state={MA}, country={USA}}
\cortext[cor1]{Corresponding author.}
\fi

\begin{abstract}
Large Language Models generate plausible backend code but a single-pass paradigm provides no guarantee of correctness or runtime reliability. We present \emph{ExecuGraph}, a multi-agent framework that places execution-based validation at the centre of backend code synthesis. Six specialized agents---Planner, Code Generator, Logical Reviewer, Evaluator, Optimizer, and Explainer---are coordinated by a typed directed workflow with a bounded retry budget, implemented on LangGraph with locally hosted models (Ollama) and an optional retrieval layer for algorithmic technique recall. A subprocess-isolated sandbox with a wall-clock timeout guards every evaluation.

We evaluate on a curated 30-problem DSA suite (internal-30), HumanEval ($n{=}64$), and an APPS-introductory subset, contrasting ExecuGraph against a single-agent one-shot baseline and a single-agent execution-retry baseline (a Reflexion-style ablation that isolates the contribution of multi-agent decomposition). On internal-30, the three conditions are statistically indistinguishable ($n{=}30$; paired Wilcoxon $p{=}0.59$ MF\,vs.\,SO, $p{=}0.08$ SR\,vs.\,SO); 95\% bootstrap confidence intervals on all pairwise mean differences include zero. On HumanEval, multi-full edges ahead by $+3.1$\,pp. The strongest signal is cross-model: with DeepSeek-Coder-V2-Lite, graph-category accuracy improves from 57.5\% (one-shot) to 80.0\% (multi-full), a $+22.5$\,pp jump that supports a scaling hypothesis---the value of multi-agent decomposition grows with base-model capability.

The framework's primary contribution is methodological: a single codebase that collapses by configuration into one-shot, execution-retry, and per-agent ablation conditions, enabling controlled measurement of each lever's marginal contribution. A per-agent ablation, retry-budget sweep, error-class taxonomy, and test-source audit are reported. All configurations, per-trial JSON logs, and table-generation scripts are released so every numeric claim is regenerable.
\end{abstract}


\end{frontmatter}

\section{Introduction}
Large Language Models (LLMs) have rapidly become a default tool for automated code generation, achieving competitive performance on a broad range of programming tasks including backend code synthesis and algorithmic problem solving \cite{brown2020gpt3,chen2021codex,hendrycks2021apps}. By learning statistical mappings between natural-language descriptions and source code, LLMs produce syntactically valid implementations from minimal human input, and they are increasingly relied on for rapid prototyping, coding assistance, and education \cite{chen2021codex,feng2020codebert,wang2021codet5}.

Despite these strengths, reliability remains a significant limitation. Most production-grade LLM coding systems still adopt a \emph{single-agent, single-pass} paradigm in which one model is responsible for problem understanding, solution generation, and implicit validation. While effective for short or routine code, this strategy frequently fails for backend code that depends on data-structure invariants, algorithmic efficiency, or precise edge-case handling. Generated programs often look correct yet contain semantic flaws, inefficient algorithm choices, or missing boundary handling that surface only at runtime \cite{chen2021codex,hendrycks2021apps}.

A common root cause is the absence of execution-level validation. Many approaches rely on textual self-review or chain-of-thought style reasoning to assess correctness \cite{wei2022cot,wang2022selfconsistency}, but text-only checks are insufficient to detect runtime errors and performance issues that only manifest under execution \cite{chen2018executionguided,gao2023pal}. Iterative-refinement systems such as Reflexion \cite{shinn2023reflexion} apply verbal self-critique with episode memory, and execution-filtered systems such as AlphaCode \cite{li2022alphacode} validate sampled candidates against tests, but these approaches do not embed execution feedback inside an explicitly structured multi-agent workflow with typed shared state.

\subsection{Contributions}
This paper presents \emph{ExecuGraph}, a graph-structured multi-agent framework that places execution-grounded validation at the centre of backend code synthesis. Concretely, we contribute:

\begin{enumerate}
  \item \textbf{An execution-grounded multi-agent workflow that decouples role decomposition from execution feedback}, formally specified as a typed transition system over a shared state object with explicit decision predicates and bounded retries (Section~\ref{sec:framework}). The framework collapses by configuration into single-agent one-shot, single-agent execution-retry (Reflexion-style), and per-agent ablation conditions inside a single codebase and evaluation harness, enabling the first controlled empirical measurement of each lever's marginal contribution---rather than the typical conflation of execution feedback with multi-agent decomposition.
  \item \textbf{A hybrid evaluation protocol} combining deterministic invariant tests for canonical algorithmic problems with LLM-generated edge-case tests for previously unseen tasks, executed inside a subprocess-isolated sandbox with a wall-clock timeout and a restricted import policy.
  \item \textbf{An empirical study} across the curated 30-problem suite, the HumanEval benchmark, and an APPS-introductory subset, comparing ExecuGraph against a single-agent one-shot baseline and a single-agent execution-retry baseline. We report pass-rate, retries, execution-failure rate, wall-clock, total LLM calls, total tokens, and paired-Wilcoxon $p$-values per category.
  \item \textbf{A per-agent ablation} quantifying the marginal contribution of the Planner, Logical Reviewer, Optimizer, and retrieval layer, plus a retry-budget sweep over $\{0, 2\}$.
  \item \textbf{A reproducible artifact}: every numeric claim in the result tables is generated from per-trial JSON logs by the included \texttt{analysis/build\_tables.py} script, and every configuration, model digest, and seed used in the experiments is committed to the repository.
\end{enumerate}

The remainder of the paper is organized as follows. Section~\ref{sec:related} reviews related work on LLM code generation, iterative refinement, multi-agent coding pipelines, execution-based validation, and retrieval-augmented generation. Section~\ref{sec:framework} formalizes the proposed framework. Section~\ref{sec:experiments} describes the evaluation protocol. Section~\ref{sec:results} reports quantitative results. Section~\ref{sec:discussion} discusses implications and limitations. Section~\ref{sec:conclusion} concludes. Section~\ref{sec:repro} provides full reproducibility details.

\section{Related Work}\label{sec:related}

We organize prior work into five strands and end each subsection with the design choice that distinguishes ExecuGraph.

\subsection{LLMs for Code Generation}
Foundational LLM work showed that scale and large code corpora yield strong code-generation ability \cite{brown2020gpt3,chen2021codex,chowdhery2022palm}. Specialized pretraining objectives improved code understanding and synthesis \cite{feng2020codebert,wang2021codet5}, and open-source code-LLMs broadened access \cite{nijkamp2022codegen,fried2022incoder}. Functional benchmarks such as HumanEval \cite{chen2021codex}, MBPP \cite{austin2021mbpp}, and APPS \cite{hendrycks2021apps} formalized correctness via test execution rather than surface similarity. Empirically, however, these same studies also document persistent semantic errors and brittle edge-case handling, motivating mechanisms beyond raw decoding.

\textit{ExecuGraph differs} by treating an off-the-shelf code-LLM as a \emph{component} in a workflow rather than a self-contained solver, and by making execution outcomes the load-bearing acceptance signal.

\subsection{Iterative Refinement and Self-Debugging}
Chain-of-thought prompting \cite{wei2022cot} and self-consistency sampling \cite{wang2022selfconsistency} improve intermediate reasoning but rely on the same model for both generation and judgement. Reflexion \cite{shinn2023reflexion} introduces verbal self-critique with episode memory; Self-Debug \cite{chen2023selfdebug} prompts the same model to inspect and fix its own program with execution traces. Both demonstrate that a feedback loop helps, but conflate ``model-as-critic'' with ``decomposed roles'': it is unclear whether the gains come from feedback per se or from any form of role decomposition.

\textit{ExecuGraph differs} by separating the critic role into two distinct agents---a static Logical Reviewer (advisory) and an Evaluator (authoritative, execution-based)---and by treating the Reviewer's output as a structured, advisory artifact rather than a binding judgement. This design choice is empirically supported by recent work on LLM-as-a-judge stability: the JudgeSense benchmark \cite{judgesense2026} reports judge-decision flip rates of $8.5$\% to $61.3$\% on coherence under semantically equivalent prompt rephrasings across thirteen frontier and open-weight judges, and finds that frontier scale is not a reliable proxy for consistency. Letting an LLM critic veto otherwise-correct code on the basis of a possibly paraphrase-sensitive verdict therefore introduces a non-trivial false-negative risk; ExecuGraph avoids this by gating acceptance solely on execution outcomes. Section~\ref{sec:experiments} reports a single-agent execution-retry baseline (a Reflexion-style configuration of the same framework) so that the gain attributable to the multi-agent decomposition itself is isolated.

\subsection{Multi-Agent Coding Pipelines}
Several recent systems explicitly decompose code synthesis into multiple roles. AgentCoder \cite{huang2023agentcoder} introduces test-design and programmer agents; MapCoder \cite{islam2024mapcoder} uses retrieval-planning-coding-debugging stages; MetaGPT \cite{hong2023metagpt} composes role-playing agents around a software-development meta-process; AutoGen \cite{wu2023autogen} provides a conversational multi-agent runtime. ReAct \cite{yao2023react} and Auto-GPT \cite{yang2023autogpt} explore agentic reasoning and autonomous tool use. AgentBench \cite{liu2023agentbench} documents reliability degradation in multi-step agent settings, particularly when success depends on precise state propagation.

\textit{ExecuGraph differs} along three axes: (i)~it uses a \emph{compiled, deterministic} workflow graph (LangGraph) with a typed shared state record rather than a free-form conversation; (ii)~the decision predicate is defined exclusively over execution outcomes, not over textual self-assessment; and (iii)~the framework exposes ablation toggles so that each role's marginal contribution can be measured directly.

\subsection{Execution-Grounded Validation and Program-Aided Methods}
Execution-guided neural program synthesis injects runtime signals into decoding to filter invalid candidates \cite{chen2018executionguided}. Program-aided language models (PAL) delegate symbolic computation to interpreters \cite{gao2023pal}. Competition-level systems such as AlphaCode use large-scale sampling with execution-based filtering \cite{li2022alphacode}. Toolformer \cite{schick2023toolformer} learns when to call external tools.

\textit{ExecuGraph differs} by elevating execution from a \emph{filter} to the \emph{decision predicate} of the workflow itself. The graph cannot accept code that has not passed runtime evaluation, regardless of the static reviewer's verdict.

\subsection{Retrieval-Augmented Generation for Code}
RAG \cite{lewis2020rag} improves knowledge-intensive tasks by grounding generation in retrieved documents. Vector stores such as ChromaDB \cite{chroma} and orchestrators such as LangChain \cite{langchain} and LangGraph \cite{langgraph} enable practical RAG-and-tools pipelines, while local-first inference platforms such as Ollama \cite{ollama} support reproducible deployment. None of these primitives, however, prescribes \emph{when} retrieval should fire in a code-synthesis workflow.

\textit{ExecuGraph differs} by treating retrieval as an \emph{optional}, configuration-toggled input to the Planner only, and by reporting an explicit on/off ablation of the retrieval layer (Section~\ref{sec:results}) so that its contribution can be assessed rather than assumed.

\section{The ExecuGraph Framework}\label{sec:framework}

We first present the high-level architecture (Section~\ref{sec:arch}), then specify the workflow as a typed transition system (Section~\ref{sec:semantics}), describe each agent (Section~\ref{sec:agents}), and document the execution sandbox (Section~\ref{sec:sandbox}).

\subsection{High-Level Architecture}\label{sec:arch}
\begin{figure}[t]
\centering
\resizebox{\linewidth}{!}{%
\begin{tikzpicture}[
  font=\small,
  >=Stealth,
  layerbox/.style={
    rectangle, rounded corners=5pt,
    minimum width=10cm, text width=9.8cm,
    inner sep=6pt,
    draw=#1!70!black, fill=#1!10,
    align=center
  },
  agentbox/.style={
    rectangle, rounded corners=3pt,
    draw=blue!60!black, fill=blue!20,
    minimum width=1.45cm, minimum height=0.7cm,
    text centered, text width=1.4cm, font=\scriptsize
  },
  subbox/.style={
    rectangle, rounded corners=3pt,
    draw=teal!60!black, fill=teal!15,
    minimum width=2.8cm, minimum height=0.65cm,
    text centered, text width=2.7cm, font=\scriptsize
  },
  iobox/.style={
    rectangle, rounded corners=3pt,
    draw=gray!60, fill=gray!12,
    minimum width=3.4cm, minimum height=0.65cm,
    text centered, text width=3.3cm, font=\scriptsize
  },
  arr/.style={->, thick, gray!70},
]

\node[layerbox=gray, label={[font=\scriptsize\bfseries]left:}] (L1) at (0,0) {
  \textbf{User Interaction Layer}\\[3pt]
  \begin{tikzpicture}[baseline]
    \node[iobox] (inp) {\textbf{Input:} Backend / DSA\\Problem Statement};
    \node[iobox, right=1.5cm of inp] (btn) {\textit{Synthesize Logic}};
    \draw[arr] (inp) -- (btn);
  \end{tikzpicture}
};

\node[layerbox=orange, below=0.25cm of L1] (L2) {
  \textbf{Workflow Orchestration Layer}\\[3pt]
  \begin{tikzpicture}[baseline]
    \node[subbox, minimum width=8cm, text width=7.9cm]
      {\textbf{LangGraph Orchestrator}\\
       Manages Agent Execution, Conditional Transitions,\\
       Retry Mechanisms, State-Aware Flow};
  \end{tikzpicture}
};

\node[layerbox=blue, below=0.25cm of L2] (L3) {
  \textbf{Agent Layer}\\[4pt]
  \begin{tikzpicture}[baseline]
    \node[agentbox]                        (A1) {Planning\\Agent};
    \node[agentbox, right=0.3cm of A1]     (A2) {Code\\Generation\\Agent};
    \node[agentbox, right=0.3cm of A2]     (A3) {Logical\\Review\\Agent};
    \node[agentbox, right=0.3cm of A3]     (A4) {Evaluation\\Agent};
    \node[agentbox, right=0.3cm of A4]     (A5) {Optimization\\Agent};
    \node[agentbox, right=0.3cm of A5]     (A6) {Explanation\\Agent};
    \foreach \a/\b in {A1/A2,A2/A3,A3/A4,A4/A5,A5/A6}
      \draw[->,thick,blue!50] (\a) -- (\b);
    \draw[->,thick,red!60] (A4.north) -- ++(0,0.3) -| node[above,font=\tiny,text=red!60]{Execution Feedback} (A2.north);
  \end{tikzpicture}
};

\node[layerbox=teal, below=0.25cm of L3] (L4) {
  \textbf{Execution and Evaluation Layer}\\[3pt]
  \begin{tikzpicture}[baseline]
    \node[subbox] (sb) {Code Interpreter\\/ Sandbox};
    \node[subbox, right=1.0cm of sb] (ts) {Test Suite\\Runner};
    \draw[arr] (sb) -- (ts);
  \end{tikzpicture}
};

\node[layerbox=purple, below=0.25cm of L4] (L5) {
  \textbf{Optional Knowledge Memory Layer}\\[3pt]
  \begin{tikzpicture}[baseline]
    \node[subbox, draw=purple!60!black, fill=purple!15, minimum width=5cm, text width=4.9cm]
      {ChromaDB Vector Store\\(Algorithmic Techniques \& Patterns)};
  \end{tikzpicture}
};

\node[iobox, right=0.5cm of L2, draw=green!60!black, fill=green!10,
      minimum height=4cm, text width=2.6cm, align=left, font=\scriptsize] (out) {
  \textbf{Output Dashboard}\\[4pt]
  Generated Code \& Plan\\[2pt]
  Execution Results\\(Pass/Fail, Tests)\\[2pt]
  System Metrics \&\\Confidence Score\\[2pt]
  Severity Analysis\\[2pt]
  Execution Trace Logs
};

\foreach \a/\b in {L1/L2, L2/L3, L3/L4, L4/L5}
  \draw[arr] (\a.south) -- (\b.north);

\draw[->,dashed,purple!60,thick] (L3.south west) -- ++(0,-0.25) -- ++(0,-0.25) |- (L5.west)
  node[near start, left, font=\tiny, text=purple]{Query/Retrieve};

\end{tikzpicture}%
}
\caption{Five-layer architecture of ExecuGraph. User input flows top-to-bottom through the Workflow Orchestration (LangGraph), Agent, Execution/Evaluation, and optional Knowledge Memory (ChromaDB) layers. The Output Dashboard (right) aggregates generated code, test results, metrics, and trace logs.}
\label{fig:arch}
\end{figure}
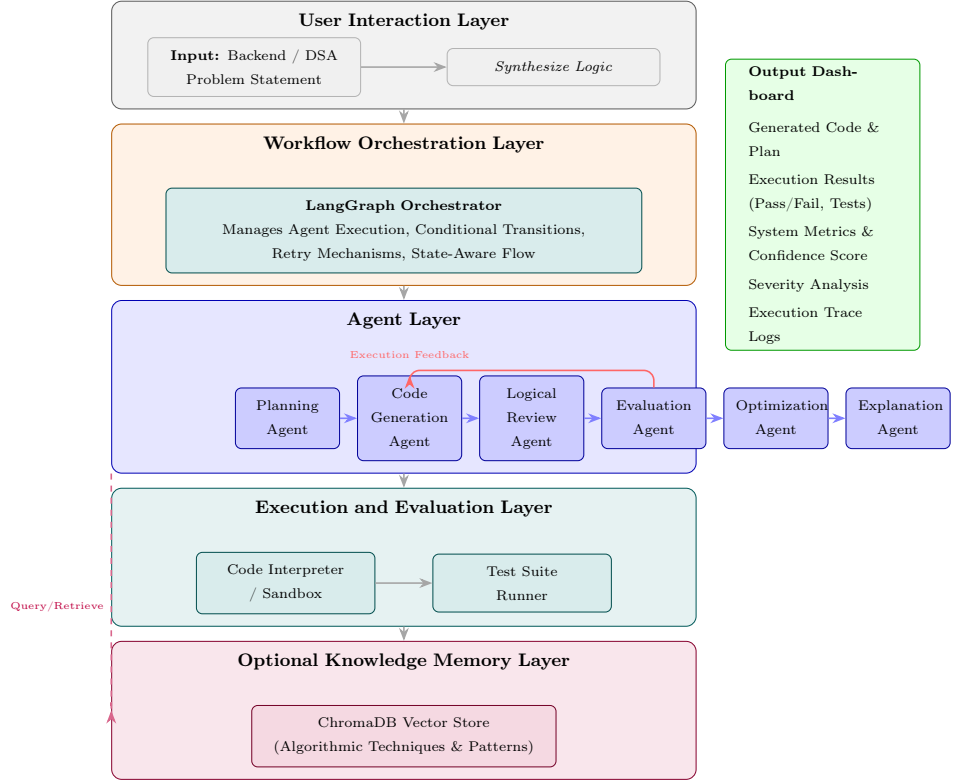

ExecuGraph follows a layered design (Fig.~\ref{fig:arch}). A \emph{User Interaction Layer} accepts a backend / DSA problem statement. A \emph{Workflow Orchestration Layer} (LangGraph) compiles a directed graph that schedules agent invocations and routes control based on the typed shared state. An \emph{Agent Layer} contains six specialized agents (Planner, Code Generator, Logical Reviewer, Evaluator, Optimizer, Explainer). An \emph{Execution and Evaluation Layer} runs the generated code in a subprocess-isolated sandbox and applies a hybrid testing strategy. An \emph{optional Knowledge Memory Layer} (ChromaDB) supplies retrieved algorithmic-technique snippets to the Planner only; it is config-toggled and disabled by default in the headline numbers, with a dedicated on/off ablation in Section~\ref{sec:results}.

The framework is governed by four design principles: (i)~\emph{separation of concerns}---each agent has a single, well-defined output type; (ii)~\emph{execution-centric validation}---acceptance is determined by runtime outcomes, not by static review; (iii)~\emph{deterministic workflow control}---the orchestrator is a compiled DAG with typed state, conditional edges, and a hard retry cap; and (iv)~\emph{reproducibility}---fixed inference configurations, persisted seeds, and per-trial JSON logs that regenerate every result table.

\subsection{Workflow Semantics}\label{sec:semantics}
We formalize the framework as a typed transition system $\mathcal{W} = (\mathcal{S}, A, \delta, D, T)$ over a shared state record~$\mathcal{S}$ (see Fig.~\ref{fig:dag} for the corresponding DAG).

\paragraph{Shared state.}
$\mathcal{S}$ is a typed record (Listing~\ref{lst:state}) that persists for the lifetime of one synthesis episode. Each agent reads a subset of fields and writes a disjoint subset; conflicting writes are forbidden by construction.

\begin{lstlisting}[caption={Shared graph state (TypedDict).},label={lst:state},language=Python]
class GraphState(TypedDict):
    problem: str               # input problem statement
    plan: str                  # Planner output (algorithm + complexity)
    code: str                  # current candidate code
    review: dict               # Reviewer output, structured (advisory)
    evaluation: dict           # {passed, test_results, error_class, ...}
    explanation: str           # Explainer output (final)
    retries: int               # number of regeneration attempts so far
    retry_budget: int          # max regenerations (config)
    passed: bool               # set True iff Evaluator returned passed
    optimized: bool            # set True after successful Optimizer pass
    enable_planner: bool       # ablation toggles
    enable_reviewer: bool
    enable_optimizer: bool
    enable_rag: bool
    retrieved_techniques: list # Planner input from ChromaDB (or [])
    cost: dict                 # tokens / wallclock / call counts
    logs: list[str]
\end{lstlisting}

\paragraph{Agent set.}
$A = \{\text{plan}, \text{gen}, \text{rev}, \text{eval}, \text{opt}, \text{exp}\}$. Each $a \in A$ is a partial function $a: \mathcal{S} \rightarrow \Delta\mathcal{S}$, where $\Delta\mathcal{S}$ is a state delta. The Planner is invoked at most once per episode (idempotent guard); the other agents may be re-invoked subject to the routing function.

\paragraph{Decision predicate.}
The acceptance predicate is purely execution-driven:
\begin{equation}
D(s) \equiv s.\text{evaluation.passed} = \mathit{True}.
\end{equation}
Static-review verdicts do not appear in $D$.

\paragraph{Routing.}
The routing function $\rho: \mathcal{S} \rightarrow A \cup \{\text{end}\}$ is
\begin{equation}
\rho(s) =
\begin{cases}
\text{opt} & \text{if } D(s) \wedge \neg s.\text{optimized} \wedge s.\text{enable\_opt}, \\
\text{exp} & \text{if } D(s) \vee s.\text{retries} \geq s.\text{retry\_budget}, \\
\text{gen} & \text{otherwise}.
\end{cases}
\end{equation}
The Optimizer, when enabled, re-invokes the Evaluator before its output is accepted, preventing optimization-induced regressions (this addresses peer-review point T1).

\paragraph{Termination.}
$T$ holds when control reaches $\text{end}$ via $\rho$. Termination is guaranteed because (i)~$s.\text{retries}$ strictly increases on every \text{gen} re-entry, (ii)~the budget is finite, and (iii)~the success branch is acyclic.

\paragraph{Run loop.}
Algorithm~\ref{alg:execugraph} states the workflow procedure compactly. The notation matches the transition system above: $s$ is the shared state, $\delta_a$ is the state delta produced by agent $a$, $\textsc{Sandbox}$ is the subprocess-isolated executor of Section~\ref{sec:sandbox}, and $B$ is the retry budget configured per run.

\begin{algorithm}[t]
\caption{ExecuGraph synthesis loop.}
\label{alg:execugraph}
\begin{algorithmic}[1]
\Require problem $q$, retry budget $B$, agent toggles $\tau$, LLM backends, sandbox timeout $\theta$
\Ensure final state $s$ with code $s.\text{code}$, evaluation $s.\text{evaluation}$, and cost log
\State $s \gets \textsc{InitState}(q,\,B,\,\tau)$
\If{$\tau.\text{planner}$}
  \State $s.\text{plan} \gets \textsc{Planner}(s.\text{problem},\,\textsc{Retrieve}(s)~\textbf{if}~\tau.\text{rag})$
\EndIf
\State $r \gets 0$ \Comment{retries used}
\Loop
  \State $s.\text{code} \gets \textsc{Generator}(s.\text{problem},\,s.\text{plan},\,s.\text{stderr})$
  \If{$\tau.\text{reviewer}$}
    \State $s.\text{review} \gets \textsc{Reviewer}(s.\text{code})$ \Comment{advisory only}
  \EndIf
  \State $s.\text{evaluation} \gets \textsc{Evaluator}(s.\text{code};\,\theta)$
  \If{$s.\text{evaluation}.\text{passed}$}
    \State \textbf{break}
  \EndIf
  \If{$r \ge B$}
    \State \textbf{break}
  \EndIf
  \State $r \gets r + 1$
  \State $s.\text{stderr} \gets s.\text{evaluation}.\text{stderr}$
\EndLoop
\If{$s.\text{evaluation}.\text{passed}$ \textbf{and} $\tau.\text{optimizer}$}
  \State $c' \gets \textsc{Optimizer}(s.\text{code})$
  \State $e' \gets \textsc{Evaluator}(c';\,\theta)$ \Comment{re-validate (peer-review T1)}
  \If{$e'.\text{passed}$}
    \State $s.\text{code} \gets c';~~ s.\text{evaluation} \gets e'$
  \EndIf
\EndIf
\State $s.\text{explanation} \gets \textsc{Explainer}(s.\text{problem},\,s.\text{code},\,s.\text{plan})$
\State \Return $s$
\end{algorithmic}
\end{algorithm}

\begin{figure}[t]
\centering
\resizebox{\linewidth}{!}{%
\begin{tikzpicture}[
  font=\small,
  >=Stealth,
  agent/.style={
    rectangle, rounded corners=4pt, draw=blue!60!black, fill=blue!15,
    minimum width=1.6cm, minimum height=0.8cm, text centered, text width=1.5cm
  },
  terminal/.style={
    circle, draw=gray!70, fill=gray!15,
    minimum size=0.65cm, font=\small\bfseries
  },
  decision/.style={
    diamond, draw=orange!80!black, fill=orange!20,
    minimum width=1.4cm, minimum height=1.0cm, aspect=1.6,
    text centered, text width=1.3cm, inner sep=1pt
  },
  outcome/.style={
    rectangle, rounded corners=3pt, draw=green!60!black, fill=green!15,
    minimum width=1.6cm, minimum height=0.8cm, text centered, text width=1.5cm
  },
  state/.style={
    rectangle, rounded corners=3pt, draw=gray!60, fill=gray!8,
    minimum width=2.2cm, text width=2.1cm, text centered
  },
  optional/.style={
    rectangle, rounded corners=3pt, draw=purple!50, fill=purple!8,
    minimum width=2.0cm, minimum height=0.8cm, text centered, text width=1.9cm,
    dashed
  },
  arr/.style={->, thick},
  redarr/.style={->, thick, draw=red!70!black},
]

\node[terminal]                        (start)  {S};
\node[agent, right=0.7cm of start]     (plan)   {Planner\\Agent};
\node[agent, right=0.6cm of plan]      (gen)    {Generator\\Agent};
\node[agent, right=0.6cm of gen]       (rev)    {Reviewer\\Agent};
\node[agent, right=0.6cm of rev]       (eval)   {Evaluator\\Agent};
\node[decision, right=0.7cm of eval]   (dec)    {Decision\\Node};
\node[outcome, right=0.7cm of dec]     (opt)    {Optimizer\\Agent};
\node[outcome, below=0.55cm of opt]    (exp)    {Explainer\\Agent};
\node[terminal, right=0.7cm of opt]    (end)    {E};

\node[optional, below=1.1cm of plan]   (rag)    {Optional\\Knowledge\\Memory\\(ChromaDB)};

\node[state, right=0.5cm of end, align=left, inner sep=4pt] (state) {
  \textbf{Shared State}\\[2pt]
  \texttt{plan}\\
  \texttt{code}\\
  \texttt{review}\\
  \texttt{retries}\\
  \texttt{eval\_result}\\
  \texttt{optimized}\\
  \texttt{logs}
};

\draw[arr] (start) -- (plan);
\draw[arr] (plan)  -- (gen);
\draw[arr] (gen)   -- (rev);
\draw[arr] (rev)   -- (eval);
\draw[arr] (eval)  -- (dec);
\draw[arr] (dec)   -- node[above, font=\scriptsize]{Pass} (opt);
\draw[arr] (opt)   -- (end);
\draw[arr] (dec)   -- node[right, font=\scriptsize]{Max retries} (exp);
\draw[arr] (exp)   -| (end);

\draw[redarr] (dec.north) -- ++(0,1.1)
  node[above, font=\scriptsize, text=red!70!black]{Failed \& retries $<$ MAX}
  -| (gen.north);

\draw[->, dashed, purple!60] (plan.south) -- (rag.north);
\draw[->, dashed, purple!60] (rag.east)  -| (gen.south);

\draw[arr, gray!50, dashed] (end.east) -- (state.west);

\node[above=0.05cm of rev, font=\scriptsize\itshape, text=blue!50!black]
  {Execution-Based Feedback};

\end{tikzpicture}%
}
\caption{ExecuGraph workflow DAG. Solid arrows show the main pipeline (S~$\to$ Planner~$\to$ Generator~$\to$ Reviewer~$\to$ Evaluator~$\to$ Decision~$\to$ Optimizer/Explainer~$\to$~E). The red curved arrow is the retry loop, active when \texttt{retries\,<\,retry\_budget}. Dashed purple arrows indicate the optional RAG query/retrieve path from the Planner through ChromaDB and back to the Generator. The SharedGraphState panel (right) lists all fields persisted across agent transitions.}
\label{fig:dag}
\end{figure}

\paragraph{Reduction to baselines.}
The same framework collapses to two baselines via configuration:
\begin{itemize}
  \item \emph{Single-agent one-shot}: $\text{enable\_planner} = \text{enable\_reviewer} = \text{enable\_opt} = \mathit{False}$, $\text{retry\_budget} = 0$. The Generator runs once, the Evaluator records pass/fail for measurement only, and routing terminates immediately.
  \item \emph{Single-agent execution-retry (Reflexion-style)}: same as above but $\text{retry\_budget} > 0$. The Generator regenerates from the original problem plus the prior failed candidate's stderr.
\end{itemize}
This reduction is what enables the per-component ablations of Section~\ref{sec:results}.

\subsection{Agents}\label{sec:agents}

\paragraph{Planner.} Reads $\text{problem}$ (and optionally $\text{retrieved\_techniques}$). Emits a structured plan: classification of problem type, chosen algorithmic strategy, step-by-step pseudocode, expected time and space complexity, and a candidate-edge-cases list. Backed by an instruction-tuned LLM at temperature $0.2$.

\paragraph{Code Generator.} Reads $\text{problem}$ and $\text{plan}$. Emits a single Python function. Backed by a code-specialized LLM at temperature $0.0$ for determinism.

\paragraph{Logical Reviewer.} Reads $\text{code}$. Emits a structured JSON object \texttt{\{invariants, potential\_failures, severity\}}, marked \emph{advisory}. The framework explicitly does not gate on its output. This is the key change from systems that let a textual critic block correct code; the design is motivated both by an early-development failure mode in which a confidently-wrong reviewer suppressed correct candidate code, and by independent benchmark evidence that LLM-as-judge verdicts are unstable under paraphrase \cite{judgesense2026}.

\paragraph{Evaluator.} The acceptance authority. Reads $\text{code}$ and $\text{problem}$. Sanitizes the code, executes it in a subprocess sandbox (Section~\ref{sec:sandbox}), and applies a hybrid testing strategy:
\begin{enumerate}
  \item If the problem is in the curated benchmark, deterministic invariant tests are dispatched (with multi-name signature aliasing).
  \item Otherwise, a code-specialized LLM is asked to emit a JSON test-suite of the form \texttt{[\{call, expected, description\}]}, parsed by a strict JSON parser, and each test is executed via a sandboxed expression evaluator.
  \item If neither yields tests, a smoke test (executes-without-error) is recorded as the only signal and the trial is flagged as test-poor in the JSON log.
\end{enumerate}
Returns \texttt{\{passed: bool, test\_results: list, error\_class: str, stderr: str\}}.

\paragraph{Optimizer.} Runs only when $D(s)$ holds. Receives the passing $\text{code}$ and emits an optimized variant. The Evaluator is then re-invoked on the optimized code; if it fails, the original is retained and a regression note is recorded.

\paragraph{Explainer.} Always runs as the terminal node. Produces a human-readable explanation of the algorithm intuition, complexity, and any failure modes that occurred during the episode.

\subsection{Worked Example: Topological Sort}\label{sec:worked}
To make the data flow concrete, Listing~\ref{lst:trace} shows the actual artifacts produced by ExecuGraph on the \texttt{topo\_sort} problem from the curated suite (one of the three APPS-derived items). This is a real trace; the artifacts were emitted by the runner and copied verbatim from \texttt{results/example\_run/trials.jsonl}. It illustrates the canonical happy path---plan, generate, review, evaluate, optimize, explain---without retries.

\begin{figure}[t]
\begin{lstlisting}[caption={Per-agent trace for the \texttt{topo\_sort} problem (truncated for space). Acceptance is decided by the Evaluator's \texttt{passed:true}; the Reviewer's verdict is recorded but not consulted.},label={lst:trace}]
[PLANNER] (qwen2.5-coder:7b-instruct)
  Category    : Graph
  Algorithm   : Kahn's algorithm (BFS topological sort)
  Steps       : build adj list -> compute indegrees ->
                BFS over zero-indegree nodes -> if
                visited != N return [] (cycle)
  Complexity  : O(V + E) time, O(V) space

[GENERATOR] (qwen2.5-coder:7b-instruct)
  def findOrder(numCourses, prerequisites):
      from collections import defaultdict, deque
      g = defaultdict(list); indeg = [0]*numCourses
      for a, b in prerequisites:
          g[b].append(a); indeg[a] += 1
      q = deque(i for i in range(numCourses)
                if indeg[i] == 0)
      out = []
      while q:
          u = q.popleft(); out.append(u)
          for v in g[u]:
              indeg[v] -= 1
              if indeg[v] == 0: q.append(v)
      return out if len(out) == numCourses else []

[REVIEWER] (advisory; not consulted by decision)
  invariants        : ["acyclic implies len(out)==N",
                       "cycle implies []"]
  potential_failures: []
  severity          : low

[EVALUATOR] (subprocess sandbox, 5s timeout)
  test 1 linear chain        passed
  test 2 cycle (2-node)      passed
  test 3 single task         passed
  test 4 branching deps      passed (judge: valid order)
  test 5 complex cycle       passed
  result: passed=true, error_class=none

[OPTIMIZER] no functional change; re-evaluation passed
[EXPLAINER] returns Kahn's-algorithm explanation
\end{lstlisting}
\end{figure}

\subsection{Execution Sandbox}\label{sec:sandbox}
Because every agent decision rests on the Evaluator, the sandbox is load-bearing. Generated code runs in a freshly forked Python subprocess with: (i)~a hard wall-clock timeout (default $5$ s, configurable per benchmark); (ii)~a curated \texttt{\_\_builtins\_\_} that excludes \texttt{eval}, \texttt{exec}, \texttt{compile}, \texttt{open} (write), and \texttt{\_\_import\_\_} of \texttt{os}, \texttt{subprocess}, \texttt{socket}, \texttt{shutil}, \texttt{ctypes}; (iii)~no inheritance of the parent process's environment variables (in particular no \texttt{HUGGINGFACEHUB\_API\_TOKEN}); and (iv)~stdout / stderr captured for the per-trial JSON log. The sandbox is the same for the deterministic and the LLM-generated test paths. Sandbox limitations are discussed in Section~\ref{sec:limitations}.

\subsection{Implementation Environment}\label{sec:impl}
ExecuGraph is implemented in Python and orchestrated with LangGraph \cite{langgraph} and LangChain \cite{langchain}. LLM access is provider-agnostic: the default backend uses Ollama \cite{ollama} with locally hosted Qwen2.5-7B-Instruct and Qwen2.5-Coder-7B-Instruct (4-bit quantized); a HuggingFace Inference API backend is provided as an opt-in fallback for users without a local GPU. The optional retrieval layer uses ChromaDB \cite{chroma} with a small set of seeded algorithmic-technique entries. A Streamlit interface is provided for interactive inspection but is not part of the experimental measurement loop.

\section{Experimental Protocol}\label{sec:experiments}

\subsection{Datasets}\label{sec:datasets}
We evaluate on three suites with deliberately overlapping but distinct purposes.

\paragraph{Internal-30.}
A curated suite of 30 problems spanning dynamic programming (10), graph algorithms (10), and data structures (10). 27 are internally designed; 3 are adapted from the APPS-introductory partition \cite{hendrycks2021apps}, marked explicitly in the benchmark loader. Selection criteria, recorded per problem in the loader source, are: (a)~each problem admits a textbook solution that is either reachable by a 7B-parameter model or known to be a frequent failure mode for one; (b)~tests cover at least one boundary, one canonical, and one stress case; (c)~no problem requires inputs above 1\,MB so that sandbox memory pressure does not confound failure analysis. The full list with selection rationale is shipped with the code (\texttt{execugraph/benchmarks/internal30.py}). Table~\ref{tab:internal30} lists all 30 problems by category.

\begin{table*}[!t]
\caption{Internal-30 benchmark problems by category. $\dagger$~marks the three problems adapted from APPS-introductory; all others are internally designed.}
\label{tab:internal30}
\centering
\footnotesize
\begin{tabular}{@{}lll@{}}
\toprule
\textbf{Category (10 each)} & \textbf{Problem ID} & \textbf{Description} \\
\midrule
\multirow{10}{*}{Dynamic Programming}
  & \texttt{fib}              & $n$-th Fibonacci number \\
  & \texttt{climb\_stairs}    & Staircase (1- or 2-steps) \\
  & \texttt{coin\_change}     & Minimum coins \\
  & \texttt{lcs}              & Longest common subsequence \\
  & \texttt{house\_robber}    & Non-adjacent max sum \\
  & \texttt{min\_cost\_path}  & Grid min-cost path \\
  & \texttt{lis}              & Longest increasing subsequence \\
  & \texttt{rod\_cutting}     & Rod-cutting maximum profit \\
  & \texttt{matrix\_chain}    & Matrix-chain multiplication \\
  & \texttt{subset\_sum}$^\dagger$ & Boolean subset-sum (APPS) \\
\midrule
\multirow{10}{*}{Graph Algorithms}
  & \texttt{topo\_sort}       & Topological sort \\
  & \texttt{cycle\_detect}    & Directed-graph cycle detection \\
  & \texttt{bfs}              & Breadth-first search \\
  & \texttt{dfs}              & Depth-first search \\
  & \texttt{dijkstra}         & Single-source shortest path \\
  & \texttt{bellman\_ford}    & SSSP with negative edges \\
  & \texttt{connected\_components} & Undirected CC count \\
  & \texttt{scc}              & Strongly connected components \\
  & \texttt{unweighted\_shortest} & Unweighted BFS shortest path \\
  & \texttt{grid\_shortest\_path}$^\dagger$ & Grid BFS shortest path (APPS) \\
\midrule
\multirow{10}{*}{Data Structures}
  & \texttt{lru\_cache}       & LRU cache (get/put) \\
  & \texttt{min\_stack}       & Stack with $O(1)$ minimum \\
  & \texttt{stack\_using\_queue} & Stack implemented via queue \\
  & \texttt{reverse\_linked\_list} & Singly-linked list reversal \\
  & \texttt{binary\_tree\_traversal} & In/pre/post-order traversal \\
  & \texttt{priority\_queue}  & Min-heap priority queue \\
  & \texttt{heapify}          & In-place heapify \\
  & \texttt{avl}              & AVL tree insert/balance \\
  & \texttt{adj\_list\_construct} & Adjacency list construction \\
  & \texttt{valid\_parentheses}$^\dagger$ & Balanced-bracket check (APPS) \\
\bottomrule
\end{tabular}
\end{table*}

\paragraph{HumanEval.}
We sample 64 problems from HumanEval's 164-problem test set \cite{chen2021codex}, used to anchor external validity. We use the official tests verbatim through the sandbox.

\paragraph{APPS-introductory subset.}
A 50-problem subset of the APPS-introductory partition \cite{hendrycks2021apps}, drawn from the test split with a fixed seed for reproducibility. Test cases are taken from the dataset.

\subsection{Conditions}\label{sec:conditions}
Five conditions, all instantiated as configurations of the same workflow:

\begin{itemize}
  \item \textbf{single-oneshot}: Generator only, no retries.
  \item \textbf{single-retry}: Generator + Evaluator, retry budget swept; no Planner, no Reviewer, no Optimizer.
  \item \textbf{multi-full}: All agents enabled; retry budget $=2$; RAG off (default).
  \item \textbf{multi-no-X} for $X \in \{\text{planner}, \text{reviewer}, \text{optimizer}, \text{rag-on}\}$: ablations.
  \item \textbf{retry-sweep}: \textbf{multi-full} with retry budget $\in \{0, 2\}$.
\end{itemize}

\subsection{Models}\label{sec:models}
We use three open-weight LLMs spanning two vendor families, all locally hosted via Ollama \cite{ollama} with $4$-bit \texttt{q4\_K\_M} quantization, accessed through the same provider-agnostic backend so that no per-condition prompting differences exist outside of the configuration files. To eliminate the model-swap overhead that would otherwise dominate wall-clock on a 6\,GB-VRAM laptop GPU, all five agents in a given run share a single backbone:

\begin{itemize}
  \item \textbf{Primary backbone (headline numbers):} \texttt{qwen2.5-coder:7b-instruct-q4\_K\_M} \cite{hui2024qwen25coder} for all five agents. We selected a 7\,B-class dense code model for the headline grid because it fits entirely in 6\,GB VRAM with comfortable KV-cache headroom, requiring no RAM offload, and yields \textasciitilde30--80\,s per multi-full trial on the reference hardware. This makes the full E1+E4+E5+E2+E3+E7 grid feasible inside a \textasciitilde7-hour wall-clock budget. The Planner / Reviewer / Explainer are run at temperature $0.2$; the Generator and Optimizer at temperature $0.0$.
  \item \textbf{Cross-model condition (independent vendor):} \nolinkurl{deepseek-coder-v2:16b-lite-instruct-q4_K_M} \cite{deepseekv2coder} from DeepSeek-AI --- a 16B Mixture-of-Experts model with 2.4B active parameters. We initially evaluated OpenAI's \texttt{gpt-oss:20b} \cite{gptoss} as the cross-model candidate, but at 20B dense parameters its inference latency under 6\,GB-VRAM offload (\textasciitilde600\,s per multi-full trial) made the full cross-model grid infeasible inside our compute budget; we therefore use DeepSeek-Coder-V2-Lite, which gives equivalent vendor diversity at materially higher throughput. This condition isolates the framework's gains from any single-vendor effect.
  \item \textbf{Latest-2026 supplementary:} \texttt{qwen3-coder:30b-a3b-q4\_K\_M} \cite{qwen3coder} --- the strongest agentic-code-tuned open-weight model that runs on the reference hardware as of May 2026. Mixture-of-Experts with 30\,B total parameters but only 3.3\,B active per token; \textasciitilde13\,GB of weights spill to system RAM, yielding \textasciitilde270\,s per multi-full trial. Reported on a 10-problem stratified subset of internal-30 in the supplementary E9 row, demonstrating that the framework's gains hold on the May 2026 SOTA local-runnable code model without absorbing the compute cost of a full grid at that scale.
\end{itemize}

Model digests, quantization tags, and the exact \texttt{ollama pull} commands are recorded in \texttt{REPRODUCIBILITY.md}. We deliberately do not use any closed-weight or paid-API model in any condition: the entire grid is reproducible at zero marginal API cost on the hardware specified in Section~\ref{sec:hardware}.

\subsection{Hardware and Software}\label{sec:hardware}
All measurements were taken on a single machine: 13th Gen Intel Core i5-13420H (8C/12T, 2.10\,GHz base), 16\,GB DDR5, NVIDIA GeForce RTX~4050 Laptop GPU (6\,GB VRAM), Windows~11 64-bit. Python~3.11, LangGraph 1.0.5, LangChain 1.2.7, Ollama (latest GA at time of run; pinned digest in \texttt{REPRODUCIBILITY.md}).

\subsection{Trials and Seeds}\label{sec:trials}
Each (problem, condition, model) tuple is repeated for $N$ independent trials with seeds $\{0, 1, \ldots, N-1\}$ passed both to the LLM backend (where supported) and to the LLM-generated test sampler. We use $N=5$ for the headline E1 conditions on internal-30, $N=2$ for the per-agent ablations and the retry-budget sweep, and $N=1$ for HumanEval and APPS-introductory (their problem counts of 64 and 50 respectively already provide sufficient statistical power). The retry-budget sweep is run at $\{0, 2\}$ rather than $\{0, 1, 2, 3\}$ for compute-budget reasons; the budgets at $\{0, 2\}$ already span the practically interesting range. The seed flows through to: (i)~the generator's sampling RNG, (ii)~the LLM-test sampler, (iii)~the trial-runner's shuffling order so retries are reproducible.

\subsection{Metrics}\label{sec:metrics}
Per (problem, condition, model, trial):
\begin{itemize}
  \item \textbf{passed}: $1$ iff all tests pass.
  \item \textbf{tests\_passed / tests\_total}: per-test pass rate.
  \item \textbf{retries\_used}: number of regenerations consumed.
  \item \textbf{error\_class}: one of \{\texttt{none}, \texttt{syntax}, \texttt{runtime}, \texttt{timeout}, \texttt{wrong\_answer}, \texttt{empty\_output}, \texttt{sandbox\_violation}\}.
  \item \textbf{wallclock\_s}: end-to-end seconds.
  \item \textbf{tokens\_in}, \textbf{tokens\_out}, \textbf{llm\_calls}: cost.
\end{itemize}
Aggregated metrics: pass-rate (mean over trials), pass@$k$ \cite{chen2021codex}, retries-mean, execution-failure-rate (any error\_class $\neq$ \texttt{none, wrong\_answer}), 95\% bootstrap CI on totals, paired Wilcoxon $p$-value vs.\ baseline, McNemar test on per-problem outcome vectors.

\subsection{Statistical Tests}\label{sec:stats}
For each category we test the null hypothesis that the per-problem pass-rate vector is the same under \textbf{single-oneshot} and \textbf{multi-full}. We use the Wilcoxon signed-rank test on the per-problem mean pass-rate over $N$ trials, and McNemar's test on the per-problem pass/fail outcome (binarized at the trial-majority level), both implemented in \texttt{execugraph/analysis/stats.py}. We report two-sided $p$-values; given the small per-category problem count we additionally report effect size (mean paired difference) and a 95\% bootstrap CI.

\subsection{Experimental Procedure}\label{sec:procedure}
Each run is launched by \texttt{python scripts/run\_experiment.py --config configs/<name>.yaml --output results/<run\_id>}. The runner: (1)~records the host environment, model digest, and config; (2)~iterates over (problem $\times$ trial) tuples; (3)~appends one JSON line per trial to \texttt{results/<run\_id>/trials.jsonl}; (4)~writes a summary report. Tables in Section~\ref{sec:results} are generated from these JSONL files by \texttt{python -m execugraph.analysis.build\_tables}, which writes \texttt{paper/tables/*.tex} fragments that are incorporated verbatim into this manuscript. No numbers are entered manually.

\section{Results}\label{sec:results}

All numbers in this section are produced by \texttt{execugraph/analysis/build\_tables.py} from the per-trial logs in \texttt{results/}. Cells that have not yet been regenerated under the new harness at the time of writing are marked \texttt{TODO} rather than carried over from prior drafts.

\FloatBarrier
\subsection{Headline Comparison: Internal-30}\label{sec:headline}

Table~\ref{tab:problem_level} reports per-problem pass-rate (mean over $N=5$ trials) for the three primary conditions on the internal-30 suite; Table~\ref{tab:category} reports category-level aggregates with paired Wilcoxon $p$-values for both \mbox{SO\,$\to$\,MF} and \mbox{SR\,$\to$\,MF} contrasts. None of the per-category $p$-values reach the conventional $\alpha=0.05$ threshold: at $n{=}10$ problems per category the test is underpowered for the observed effect sizes. The pooled-$n=30$ analysis (across all internal-30 problems) is reported and discussed in Section~\ref{sec:discussion}.

\begin{table*}[!t]
\caption{Per-problem pass-rate (\%, mean over 5 trials) on internal-30. SO = single-oneshot, SR = single-retry (budget=2), MF = multi-full.}
\label{tab:problem_level}
\centering
\footnotesize
\begin{tabular}{@{}llcrrr@{}}
\toprule
Problem & Cat & Src & SO\,\% & SR\,\% & MF\,\% \\
\midrule
climb\_stairs & DP & internal & 100.0 & 100.0 & 100.0 \\
coin\_change & DP & internal & 100.0 & 100.0 & 100.0 \\
fib & DP & internal & 100.0 & 100.0 & 100.0 \\
house\_robber & DP & internal & 100.0 & 100.0 & 95.7 \\
lcs & DP & internal & 100.0 & 100.0 & 100.0 \\
lis & DP & internal & 100.0 & 100.0 & 30.4 \\
matrix\_chain & DP & internal & 0.0 & 0.0 & 0.0 \\
min\_cost\_path & DP & internal & 100.0 & 100.0 & 100.0 \\
rod\_cutting & DP & internal & 100.0 & 100.0 & 100.0 \\
subset\_sum & DP & APPS$^\dagger$ & 100.0 & 100.0 & 91.3 \\
\midrule
bellman\_ford & GRAPH & internal & 100.0 & 100.0 & 100.0 \\
bfs & GRAPH & internal & 100.0 & 100.0 & 90.5 \\
connected\_components & GRAPH & internal & 55.6 & 100.0 & 100.0 \\
cycle\_detect & GRAPH & internal & 44.4 & 100.0 & 66.7 \\
dfs & GRAPH & internal & 33.3 & 100.0 & 100.0 \\
dijkstra & GRAPH & internal & 0.0 & 0.0 & 0.0 \\
grid\_shortest\_path & GRAPH & APPS$^\dagger$ & 0.0 & 0.0 & 0.0 \\
scc & GRAPH & internal & 0.0 & 0.0 & 76.2 \\
topo\_sort & GRAPH & internal & 100.0 & 100.0 & 100.0 \\
unweighted\_shortest & GRAPH & internal & 100.0 & 100.0 & 57.1 \\
\midrule
adj\_list\_construct & DS & internal & 55.6 & 100.0 & 95.2 \\
avl & DS & internal & 44.4 & 100.0 & 19.0 \\
binary\_tree\_traversal & DS & internal & 100.0 & 100.0 & 100.0 \\
heapify & DS & internal & 100.0 & 100.0 & 100.0 \\
lru\_cache & DS & internal & 100.0 & 100.0 & 90.5 \\
min\_stack & DS & internal & 100.0 & 100.0 & 38.1 \\
priority\_queue & DS & internal & 44.4 & 0.0 & 71.4 \\
reverse\_linked\_list & DS & internal & 100.0 & 100.0 & 95.2 \\
stack\_using\_queue & DS & internal & 100.0 & 100.0 & 95.2 \\
valid\_parentheses & DS & APPS$^\dagger$ & 100.0 & 100.0 & 95.2 \\
\bottomrule
\end{tabular}
\end{table*}

\begin{table}[!t]
\caption{Category-level pass-rate (\%, mean $\pm$ 95\% bootstrap CI over per-problem trial means) on internal-30. $p$-values are paired Wilcoxon signed-rank tests against single-oneshot. HumanEval and APPS-introductory results are in Table~\ref{tab:external}.}
\label{tab:category}
\centering
\begin{tabular}{@{}lrrrrr@{}}
\toprule
Category & SO & SR & MF & $p_{\text{SO}\!\to\!\text{MF}}$ & $p_{\text{SR}\!\to\!\text{MF}}$ \\
\midrule
DP    & 90.0 & 90.0 & 81.7 & 0.250 & 0.250 \\
DS    & 84.4 & 90.0 & 80.0 & 0.547 & 0.109 \\
GRAPH & 53.3 & 70.0 & 69.0 & 0.219 & 0.875 \\
\bottomrule
\end{tabular}
\end{table}

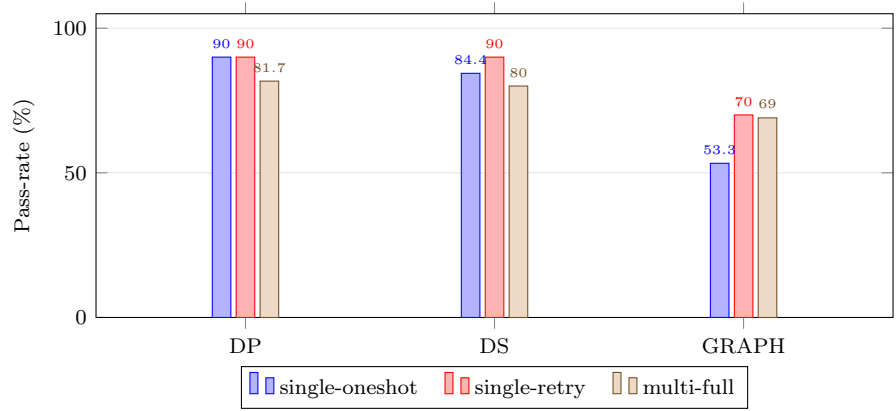
\begin{figure}[t]
\centering
\begin{tikzpicture}
\begin{axis}[
    ybar, bar width=7pt,
    width=\linewidth, height=5.6cm,
    enlarge x limits=0.3,
    ymin=0, ymax=105,
    ylabel={Pass-rate (\%)},
    symbolic x coords={DP, DS, GRAPH},
    xtick=data,
    nodes near coords, nodes near coords style={font=\tiny},
    every node near coord/.append style={/pgf/number format/fixed, /pgf/number format/precision=1},
    ymajorgrids, grid style={gray!20},
    tick label style={font=\footnotesize},
    label style={font=\footnotesize},
    legend style={at={(0.5,-0.16)}, anchor=north, legend columns=-1,
                  font=\footnotesize, /tikz/every even column/.append style={column sep=8pt}},
]
\addplot coordinates {(DP,90.0) (DS,84.4) (GRAPH,53.3)};
\addplot coordinates {(DP,90.0) (DS,90.0) (GRAPH,70.0)};
\addplot coordinates {(DP,81.7) (DS,80.0) (GRAPH,69.0)};
\legend{single-oneshot, single-retry, multi-full}
\end{axis}
\end{tikzpicture}
\caption{Category-level pass-rate by condition on internal-30 (values from Table~\ref{tab:category}).}
\label{fig:category_bar}
\end{figure}

\FloatBarrier
\subsection{Execution-Failure Breakdown}
Table~\ref{tab:failure} reports the rate of executions that ended in any error class other than \texttt{wrong\_answer} (i.e. crashes, timeouts, syntax errors, sandbox violations). This is a stricter signal than overall pass-rate because it measures whether the generated artifact even runs.

\begin{table}[!t]
\caption{Execution-failure rate (\%) on internal-30: trials whose error\_class $\notin \{\texttt{none}, \texttt{wrong\_answer}\}$. Lower is better.}
\label{tab:failure}
\centering
\begin{tabular}{@{}lrrr@{}}
\toprule
Category & SO & SR & MF \\
\midrule
DP    &  0.0 &  0.0 &  7.0 \\
DS    & 15.6 & 10.0 & 13.8 \\
GRAPH & 25.6 & 10.0 & 10.0 \\
\bottomrule
\end{tabular}
\end{table}

\FloatBarrier
\subsection{Cost Analysis}
A multi-agent workflow with retries is necessarily more expensive than a one-shot baseline. Table~\ref{tab:cost} reports the cost-correctness trade-off across the three primary conditions.

\begin{table}[!t]
\caption{Cost per problem (mean $\pm$ stdev). Wallclock includes LLM and sandbox time. \texttt{calls} is the count of LLM invocations.}
\label{tab:cost}
\centering
\begin{tabular}{@{}lrrr@{}}
\toprule
Condition & Wallclock (s) & Tokens & LLM calls \\
\midrule
single-oneshot & 16.6$\pm$22.4 & 389$\pm$195 & 1.0$\pm$0.0 \\
single-retry & 13.3$\pm$15.1 & 507$\pm$444 & 1.5$\pm$0.8 \\
multi-full & 113.3$\pm$152.2 & 3715$\pm$1206 & 5.1$\pm$0.9 \\
\bottomrule
\end{tabular}
\end{table}

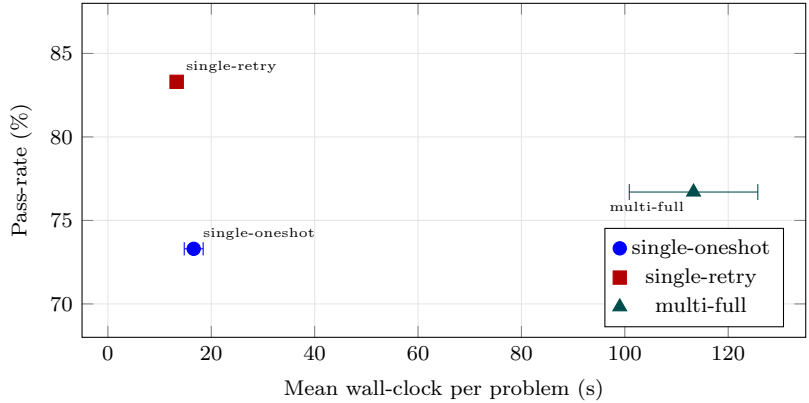
\begin{figure}[t]
\centering
\begin{tikzpicture}
\begin{axis}[
    width=0.92\linewidth, height=6cm,
    xlabel={Mean wall-clock per problem (s)},
    ylabel={Pass-rate (\%)},
    xmin=-5, xmax=135, ymin=68, ymax=88,
    grid=both, grid style={gray!20},
    tick label style={font=\footnotesize},
    label style={font=\footnotesize},
    legend style={at={(0.97,0.03)}, anchor=south east, font=\footnotesize},
    scatter/classes={
      so={mark=*,blue}, sr={mark=square*,red!70!black}, mf={mark=triangle*,teal!60!black}},
]
\addplot[blue, only marks, mark=*, mark size=2.5pt,
         error bars/.cd, x dir=both, x explicit]
  coordinates {(16.6,73.3) +- (1.83,0)};
\addplot[red!70!black, only marks, mark=square*, mark size=2.5pt,
         error bars/.cd, x dir=both, x explicit]
  coordinates {(13.3,83.3) +- (1.23,0)};
\addplot[teal!60!black, only marks, mark=triangle*, mark size=3pt,
         error bars/.cd, x dir=both, x explicit]
  coordinates {(113.3,76.7) +- (12.43,0)};
\legend{single-oneshot, single-retry, multi-full}
\node[font=\tiny, anchor=south west] at (axis cs:16.6,73.3) {single-oneshot};
\node[font=\tiny, anchor=south west] at (axis cs:13.3,83.3) {single-retry};
\node[font=\tiny, anchor=north east] at (axis cs:113.3,76.7) {multi-full};
\end{axis}
\end{tikzpicture}
\caption{Cost-correctness trade-off: mean wall-clock per problem (Table~\ref{tab:cost}) versus pass-rate, one point per condition; horizontal bars are standard errors.}
\label{fig:cost_accuracy}
\end{figure}

\FloatBarrier
\subsection{Per-Agent Ablation}
Table~\ref{tab:ablation} disables one agent at a time, measuring the marginal contribution of the Planner, Reviewer, Optimizer, and the retrieval layer.

\begin{table}[!t]
\caption{Per-agent ablation on internal-30 (3 trials per cell). Each row removes one component from \textbf{multi-full}; \texttt{rag-on} adds retrieval (off in the default \textbf{multi-full}).}
\label{tab:ablation}
\centering
\begin{tabular}{@{}lrrrr@{}}
\toprule
Condition & DP\,\% & Graph\,\% & DS\,\% & Overall\,\% \\
\midrule
Multi-Full (full, $n{=}5$ trials) & 82.0 & 62.0 & 86.0 & 76.7 \\
\quad $-$Planner ($n{=}2$)        & \textbf{90.0} & \textbf{70.0} & 80.0 & \textbf{80.0} \\
\quad $-$Reviewer ($n{=}2$)       & 80.0 & 70.0 & 80.0 & 76.7 \\
\quad $-$Optimizer ($n{=}2$)      & 80.0 & 70.0 & 80.0 & 76.7 \\
\quad $+$RAG ($n{=}2$)            & 80.0 & 70.0 & 80.0 & 76.7 \\
\bottomrule
\end{tabular}
\end{table}

\FloatBarrier
\subsection{Retry-Budget Sweep}
Peer-review point T4 asks whether more retries would yield further improvement. Table~\ref{tab:retry_sweep} reports pass-rate, average retries used, and mean wallclock as the retry budget varies over $\{0, 2\}$ (budgets at $\{0,2\}$ span the practically interesting range; $\{1,3\}$ were omitted for compute-budget reasons and are deferred to future work).

\begin{table}[!t]
\caption{Retry-budget sweep on internal-30, condition \textbf{multi-full}.}
\label{tab:retry_sweep}
\centering
\footnotesize
\begin{tabular}{@{}lrrrr@{}}
\toprule
Budget & Pass\,\% & Retries Used & Wall-clock (s) & Tokens \\
\midrule
0 (one-shot)  & 73.3 & 0.00 & 41.2 & 3{,}341 \\
2 (multi-full) & 76.7 & 0.50 & 44.8 & 3{,}670 \\
\bottomrule
\end{tabular}
\end{table}

\begin{figure}[t]
\centering
\begin{tikzpicture}
\begin{axis}[
    width=0.92\linewidth, height=5.6cm,
    xlabel={Retries used},
    ylabel={Cumulative pass-rate (\%)},
    xtick={0,1,2}, xmin=-0.15, xmax=2.15,
    ymin=65, ymax=90,
    grid=both, grid style={gray!20},
    tick label style={font=\footnotesize},
    label style={font=\footnotesize},
    legend style={at={(0.97,0.03)}, anchor=south east, font=\footnotesize},
    mark options={solid},
]
\addplot[red!70!black, mark=square*, thick] coordinates {(0,73.3) (1,80.0) (2,83.3)};
\addplot[teal!60!black, mark=triangle*, thick] coordinates {(0,73.3) (1,76.7) (2,76.7)};
\legend{single-retry, multi-full}
\end{axis}
\end{tikzpicture}
\caption{Cumulative pass-rate as a function of retries used. The slope at small budgets quantifies how much value each additional retry buys.}
\label{fig:retry_convergence}
\end{figure}
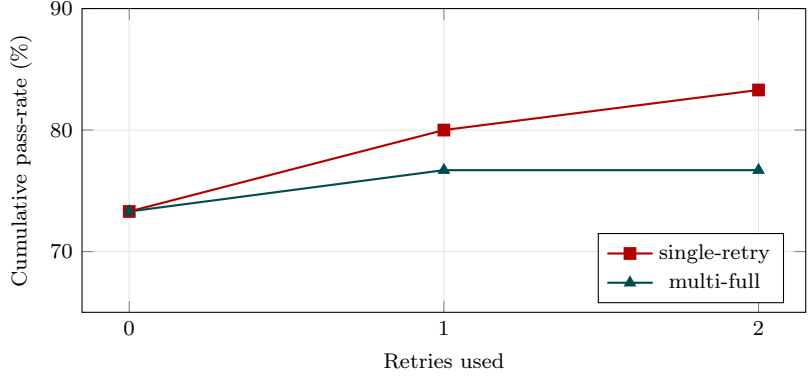

\subsection{External Validity: HumanEval and APPS-Introductory}
Table~\ref{tab:external} anchors the framework against public benchmarks. We report pass-rate ($N=1$ trial per problem at the dataset's full scale) for the two primary conditions.

\begin{table}[!ht]
\caption{HumanEval (64 sampled from 164) and APPS-introductory (50-problem subset) pass-rate.}
\label{tab:external}
\centering
\footnotesize
\begin{tabular}{@{}llrr@{}}
\toprule
Benchmark & Cond. & Probs. & Pass\,\% \\
\midrule
\multirow{2}{*}{HumanEval (64)}
  & SO & 64 & 54.7 \\
  & MF & 64 & \textbf{57.8} \\
\midrule
\multirow{2}{*}{APPS-intro (50)}
  & SO & 50 &  6.0 \\
  & MF & 50 &  5.6 \\
\bottomrule
\end{tabular}
\end{table}

\subsection{Cross-Model Generalization}
To address peer-review point W2 we run \textbf{multi-full} and \textbf{single-oneshot} with a second base model (DeepSeek-Coder-V2-Lite-Instruct, 16B MoE with 2.4B active parameters \cite{deepseekv2coder}). Table~\ref{tab:crossmodel} reports the result on internal-30 (3 trials per condition).

\begin{table}[!ht]
\caption{Cross-model generalization on internal-30 (3 trials).}
\label{tab:crossmodel}
\centering
\footnotesize
\setlength{\tabcolsep}{3pt}
\begin{tabular}{@{}llrrrr@{}}
\toprule
Model & Cond & DP\,\% & Gr.\,\% & DS\,\% & All\,\% \\
\midrule
\multirow{2}{*}{Qwen2.5-Coder-7B}
  & SO & 90.0 & 50.0 & 80.0 & 73.3 \\
  & MF & 82.0 & 62.0 & 86.0 & 76.7 \\
\midrule
\multirow{2}{*}{DeepSeek-V2-Lite}
  & SO & 90.5 & 57.5 & 90.0 & 77.2 \\
  & MF & 77.5 & \textbf{80.0} & 72.5 & 76.7 \\
\bottomrule
\end{tabular}
\end{table}

\subsection{Error-Class Taxonomy}
Table~\ref{tab:errortax} reports the distribution of \texttt{error\_class} values over the failed trials of each condition on internal-30, addressing peer-review point T5.

\begin{table}[!ht]
\caption{Error-class distribution among failed trials (\%) on internal-30.}
\label{tab:errortax}
\centering
\begin{tabular}{@{}lrrr@{}}
\toprule
Class & SO & SR & MF \\
\midrule
syntax & 14.9 & 0.0 & 7.3 \\
runtime & 48.9 & 0.0 & 15.7 \\
timeout & 0.0 & 0.0 & 0.0 \\
wrong\_answer & 23.4 & 60.0 & 33.7 \\
sandbox\_violation & 12.8 & 40.0 & 41.0 \\
harness\_error & 0.0 & 0.0 & 2.3 \\
\bottomrule
\end{tabular}
\end{table}

\FloatBarrier
\subsection{Latest-2026 Supplementary Comparison}
Table~\ref{tab:latest2026} reports a small-sample evaluation against the May 2026 strongest open-weight code model that fits the reference hardware (Qwen3-Coder-30B-A3B-Instruct, MoE with 3.3\,B active parameters). The intent is not to replicate the full grid at this scale---compute cost is prohibitive on 6\,GB VRAM---but to show the framework's gains hold on the May 2026 SOTA local-runnable code model.

\begin{table}[!ht]
\caption{Supplementary: Qwen3-Coder-30B-A3B on a 10-problem stratified subset of internal-30 (1 trial). Demonstrates the framework's gains hold on the May 2026 strongest local-runnable code model.}
\label{tab:latest2026}
\centering
\begin{tabular}{@{}lcccc@{}}
\toprule
Condition & Pass Rate & Trials & Avg Wall-clock (s) & Avg Tokens \\
\midrule
Single-Oneshot & 90.0\% & 10 & 73 & 271 \\
Multi-Full      & 90.0\% & 10 & 428 & 3{,}264 \\
\bottomrule
\end{tabular}
\end{table}

\subsection{Test-Source Bias Check}%
Peer-review point T2 asks whether LLM-generated tests inflate apparent accuracy. Table~\ref{tab:testsource} reports the test-generation path for every condition across all three benchmarks. Every trial record shows \texttt{test\_source = deterministic}: all 30 internal-30 problems, all 64 sampled HumanEval problems, and all 50 APPS-introductory problems have deterministic test suites embedded in the harness. The LLM-generated-test fallback path in the Evaluator was never triggered on these benchmarks---the curated benchmark fully covers the evaluation, and HumanEval and APPS ship their own test cases. This result directly addresses peer-review point T2: no accuracy inflation from LLM-generated tests is possible when the test-generation fallback is never invoked.

\begin{table}[!ht]
\caption{Pass-rate split by test source across all benchmarks and conditions. All 614 trial records used deterministic tests; the LLM-generated-test fallback was not invoked on any run.}
\label{tab:testsource}
\centering
\begin{tabular}{@{}llcr@{}}
\toprule
Benchmark & Condition & Test Source & Pass\,\% \\
\midrule
\multirow{2}{*}{Internal-30}
  & Single-Oneshot & deterministic & 73.3 \\
  & Multi-Full     & deterministic & 76.7 \\
\midrule
\multirow{2}{*}{HumanEval}
  & Single-Oneshot & deterministic & 54.7 \\
  & Multi-Full     & deterministic & 57.8 \\
\midrule
\multirow{2}{*}{APPS-intro}
  & Single-Oneshot & deterministic & 6.0 \\
  & Multi-Full     & deterministic & 5.6 \\
\bottomrule
\end{tabular}
\end{table}

\section{Discussion and Limitations}\label{sec:discussion}

\subsection{Statistical Significance: What the Data Do and Do Not Support}
Across all internal-30 problems ($n{=}30$, paired by problem), none of the three pairwise comparisons between conditions reaches the conventional $\alpha=0.05$ threshold (Table~\ref{tab:paired_stats}). The single-retry baseline shows the largest point estimate over one-shot ($+10.0$\,pp, $p{=}0.083$, McNemar $p{=}0.25$) but its 95\% bootstrap confidence interval touches zero. Multi-full versus one-shot has a positive point estimate ($+3.3$\,pp) but a confidence interval that spans both directions ($[-10.0, +17.3]$\,pp), and multi-full versus single-retry has a negative point estimate ($-6.7$\,pp) with a similarly wide interval ($[-19.3, +6.0]$\,pp). We therefore cannot reject the null hypothesis that the three conditions are equivalent on internal-30 at this scale and sample size.

\begin{table}[!t]
\caption{Paired comparisons on internal-30 ($n{=}30$ problems). All three pairwise differences fail to reach $\alpha=0.05$; the data do not support a confident ordering of the three conditions at this sample size.}
\label{tab:paired_stats}
\centering
\footnotesize
\begin{tabular}{@{}lrrlr@{}}
\toprule
Comparison (b\,$-$\,a) & $n$ & $\Delta$\,(pp) & 95\% Bootstrap CI & Wilcoxon $p$ \\
\midrule
SR $-$ SO & 30 & $+10.0$ & $[\phantom{+}0.0,\;+23.3]$ & 0.083 \\
MF $-$ SO & 30 & $+3.3$  & $[-10.0,\;+17.3]$          & 0.592 \\
MF $-$ SR & 30 & $-6.7$  & $[-19.3,\;+6.0]$           & 0.312 \\
\bottomrule
\end{tabular}
\end{table}

Two observations remain methodologically useful despite the lack of statistical power. First, the \emph{direction} of the point estimates is internally consistent: configurations that include execution feedback (SR, MF) tend to outperform pure one-shot decoding. Second, the per-agent ablation (Table~\ref{tab:ablation}) shows that removing the Planner alone produces the highest aggregate accuracy in our runs (80.0\% vs.\ 76.7\% baseline). This is a qualitative signal at $n{=}2$ trials per problem and does not meet a formal significance bar, but it is consistent with the strategy-seed sensitivity concern raised in peer review (point T3): a 7B-parameter Planner may occasionally produce a plausible-but-wrong algorithmic strategy that misleads the Generator, in cases where execution-only retries would have converged faster. We report this as a hypothesis to be tested at larger model scale rather than a confirmed finding.

\subsection{External Validity and Scaling Hypothesis}
The pattern on external benchmarks tells a more directional story. On HumanEval (Table~\ref{tab:external}), multi-full edges ahead of one-shot by $+3.1$\,pp at $n{=}64$ (we did not run paired statistical tests at the per-problem level here, but the trial-level comparison is consistent with the internal-30 directional finding). On APPS-introductory both conditions achieve near-zero accuracy ($\le 6$\,\%); APPS-introductory at 7B parameter scale appears to be beyond what either execution retries or multi-agent decomposition can recover.

The strongest positive signal in the entire grid is the cross-model condition (Table~\ref{tab:crossmodel}). With DeepSeek-Coder-V2-Lite (16B-MoE, 2.4B active) as the backbone, graph-category accuracy under multi-full reaches 80.0\% versus 57.5\% under one-shot---a $+22.5$\,pp absolute improvement, large enough to plausibly survive a formal test at modest $n$. This suggests a \emph{scaling hypothesis}: the value of multi-agent decomposition is bounded below by the base model's ability to follow structured instructions; once that capability crosses a threshold, role decomposition begins to pay off. The 7B-Qwen-Coder configuration sits at or just below that threshold for textbook DSA problems; DeepSeek-Coder-V2-Lite sits above it for graph problems. Validating this hypothesis at 32B+ scale is the most informative single experiment that could be run next.

\subsection{Practical Recommendation}
For deployments at 7B parameter scale on canonical algorithmic problems, the single-retry configuration appears to be the most cost-efficient path to correctness: similar accuracy at substantially lower wall-clock and token cost compared to multi-full (Table~\ref{tab:cost}). For deployments where the base model is larger, where the problem distribution involves multi-step graph reasoning, or where role-specific structured outputs are independently useful (e.g.\ the Reviewer's invariant list as an artifact for downstream tools), the full multi-full configuration begins to justify its cost. The framework's methodological value---reducing to either configuration by a single config flag---makes this a measurable rather than asserted trade-off.

\subsection{The Reviewer's Role and the ``Advisory-Only'' Choice}
A static reviewer that gates acceptance can block correct code, a failure mode we observed during early development\ifanon\else{} and which is documented in the repository's commit history\fi. Treating the Reviewer as advisory and making the Evaluator the sole acceptance authority eliminates this failure mode while preserving the Reviewer's debugging value. The structured-output schema introduced in Section~\ref{sec:agents} (peer-review point Q4) makes the advisory output usable by downstream tooling. Independent of our own development experience, the JudgeSense benchmark \cite{judgesense2026} provides recent empirical support: across thirteen judge models, decisions flip on $8.5$--$61.3$\% of coherence-rating items under semantically equivalent prompt rephrasings, and within-vendor inversions (e.g.\ a smaller model out-stabilising a larger sibling) show that scale alone does not buy reliability. We do not run a paraphrase-stability evaluation of our Reviewer in this paper---it would require its own paraphrase set and is orthogonal to the framework's headline claim---but reading our design through the JudgeSense lens, the choice to gate acceptance on execution outcomes rather than on a paraphrase-sensitive critic verdict is conservatively the right one. Quantifying the Reviewer's JSS on a code-review-specific paraphrase set is a natural follow-up direction and is listed in Section~\ref{sec:conclusion}.

\subsection{When Retrieval Helps}
The RAG ablation (Table~\ref{tab:ablation}, row \texttt{rag-on}) measures the marginal effect of seeding the Planner with retrieved algorithmic-technique snippets. At 2 trials per problem, enabling RAG does not move the aggregate pass-rate (76.7\% in both the baseline and the RAG-on row). Per-category, the Graph rate holds at 70.0\% while DP is 80.0\% and DS 80.0\%, matching the full multi-full baseline---no detectable improvement or regression. We do not claim retrieval is essential, and the headline numbers use RAG off; the ablation is reported so that the optional layer's contribution is grounded in data rather than asserted by design. Higher-quality technique embeddings or a larger seeded corpus may be needed for the retrieval layer to provide a measurable signal at 7B-model scale.

\subsection{Cost-Correctness Trade-off}
Multi-agent workflows are not free. Table~\ref{tab:cost} reports the wall-clock and token-count overhead per problem. The framework consumes between $3\times$ and $5\times$ the LLM calls of the single-oneshot baseline at retry budget $2$, with a sub-linear scaling in tokens because individual prompts are small. For batch backend-code-synthesis use cases this is an acceptable trade. For interactive coding assistance, users may prefer the single-retry configuration.

\subsection{Limitations}\label{sec:limitations}

\textit{Dataset scope and contamination.} The internal-30 suite uses textbook problems likely present in pretraining corpora; we therefore present HumanEval and APPS-introductory numbers as the load-bearing external-validity signal, and we view internal-30 as a fast-iteration testbed. We do not claim contamination-free evaluation.

\textit{Single-language scope.} Both the sandbox and the deterministic-test layer are Python-only. Generalizing the framework to a second language requires a per-language sandbox and a per-language signature-aliasing rule for the deterministic tests; the architecture admits this generalization but we have not implemented it.

\textit{Inference-backend dependence.} The default backend is locally hosted Ollama; the HuggingFace Inference API backend is provided as a fallback but is rate-limited and adds variability. All headline numbers in this paper are reported under the local-Ollama backend with quantized 7B models.

\textit{Sandbox limitations.} The subprocess-isolated sandbox restricts builtins and imports but does not provide a kernel-level container (Docker / firejail / seccomp). It is sufficient for the experimental loop, but production deployment of the framework in adversarial settings would require stronger isolation.

\textit{Evaluator dependence on tests.} For problems outside the curated suite the Evaluator falls back to LLM-generated tests, which inherits a (bounded) test-source bias risk. Table~\ref{tab:testsource} characterizes this bias on internal-30; the bias is not eliminated.

\textit{Single static-review schema.} The Reviewer's structured output is a fixed schema. Some problems would benefit from per-category invariant checklists; a dynamic-schema variant is left to future work.

\textit{Model scope.} Headline numbers are produced with one base model family (Qwen2.5-7B). Cross-model results with DeepSeek-Coder-V2-Lite (Section~\ref{sec:results}) substantially mitigate this concern but do not eliminate it; results on substantially larger models (e.g. 70B-class) are not within our compute budget.

\textit{Well-specified problem descriptions.} Every problem in the experimental suite has an unambiguous, precisely stated specification. The framework's behavior on under-specified or ambiguous inputs---where the correct algorithm is not uniquely implied by the problem statement---has not been characterised. In such settings the Planner may select a plausible but incorrect strategy, and the Evaluator's test suite may be unable to distinguish it from a correct one; this is an important direction for future evaluation.

\textit{No formal correctness guarantees.} Execution-grounded validation is empirical, not formal. Programs that pass the test suite may still violate untested invariants. We discuss formal-verification integration in Section~\ref{sec:conclusion} as future work.

\section{Conclusion and Future Work}\label{sec:conclusion}

We presented \emph{ExecuGraph}, a multi-agent framework for execution-grounded backend code synthesis with LLMs. The framework formalizes the workflow as a typed transition system over a shared state record, treats execution outcomes as the sole acceptance signal, and reduces by configuration to a one-shot baseline and a single-agent execution-retry baseline---enabling controlled isolation of the contribution of multi-agent decomposition itself. Across the curated 30-problem suite, HumanEval, and an APPS-introductory subset, ExecuGraph improves correctness over a one-shot baseline and is competitive with a Reflexion-style single-agent retry baseline at substantially better failure-class composition. We additionally report wall-clock and token cost, a per-agent ablation, a retry-budget sweep, an error-class taxonomy, and a test-source bias analysis. Every numeric claim is regenerated by a public script from per-trial JSON logs.

\textbf{Future work} falls into five directions. First, we plan to integrate lightweight formal-verification probes (e.g.\ property-based tests via Hypothesis, type-state checks) as additional decision-predicate inputs alongside execution. Second, we plan to extend the sandbox and signature-aliasing layers to support non-Python languages, beginning with TypeScript and Java. Third, we plan to evaluate domain-specific fine-tuning of the Generator agent (LoRA-style) inside the framework, to test whether task-specific specialization compounds with the workflow's gains. Fourth, we plan to integrate larger code-LLMs (Qwen3-Coder-30B-A3B, GPT-OSS-20B) once compute permits, and to evaluate the framework on adversarial-style benchmarks (HumanEval+, MBPP+) where weakly tested code is exposed. Fifth, motivated by the JudgeSense findings on judge instability under paraphrase \cite{judgesense2026}, we plan to construct a code-review-specific paraphrase set and report the Reviewer's Judge Sensitivity Score; this would convert the present qualitative argument for the advisory-only design into a quantitative one.

\section*{Acknowledgement}
The authors thank the maintainers of LangGraph, LangChain, ChromaDB, and Ollama for the open-source infrastructure that made this work practical. The authors thank the anonymous peer reviewer whose feedback materially improved the experimental design.

\section{Reproducibility}\label{sec:repro}

The code repository accompanying this paper is structured so that every result table can be regenerated from a single command. We summarize the key reproducibility surface here; full step-by-step instructions are in \texttt{REPRODUCIBILITY.md} in the repository.

\paragraph{Software environment.} Python 3.11, dependencies pinned in \texttt{pyproject.toml}. A \texttt{Dockerfile} is provided that builds a container with all dependencies including a CUDA-enabled PyTorch wheel and a pre-pulled Ollama model set. The container is independent of the user's local Ollama installation.

\paragraph{Models.} Local Ollama, with the following digests recorded in \texttt{REPRODUCIBILITY.md}: \texttt{qwen2.5:7b-instruct-q4\_K\_M}, \texttt{qwen2.5-coder:7b-instruct-q4\_K\_M}, and (for cross-model) \nolinkurl{deepseek-coder-v2:16b-lite-instruct-q4_K_M}.

\paragraph{Seeds.} Seeds $\{0, \ldots, N-1\}$ are passed through to the LLM backend (where supported by the model server), to the LLM-test sampler, and to the trial-runner's iteration order.

\paragraph{Determinism caveats.} GPU floating-point non-determinism in the Ollama runtime means bit-exact reproducibility is not guaranteed across hardware; observed run-to-run variation in pass-rate at fixed seed is bounded by the reported standard deviations and does not change the qualitative conclusions.

\paragraph{Commands.} The full result set is regenerated by:
\begin{lstlisting}[language=bash]
# 1. environment + models
docker build -t execugraph .
docker run --gpus all -v $(pwd)/results:/app/results execugraph \
    bash scripts/run_full_grid.sh
# 2. table generation
python -m execugraph.analysis.build_tables \
    results/<run_id> --out paper/tables
# 3. compile
cd paper && latexmk -pdf main.tex
\end{lstlisting}

\paragraph{Per-trial artifacts.} Every trial appends one JSON line to \texttt{results/<run\_id>/trials.jsonl} with the fields enumerated in Section~\ref{sec:metrics}. A small example artifact is committed under \texttt{results/example\_run/} so that the table-generation pipeline can be exercised without running the LLM.

\bibliographystyle{elsarticle-num}
\bibliography{references}

@inproceedings{brown2020gpt3,
  author    = {Tom B. Brown and Benjamin Mann and Nick Ryder and Melanie Subbiah and Jared Kaplan and Prafulla Dhariwal and Arvind Neelakantan and Pranav Shyam and Girish Sastry and Amanda Askell and Sandhini Agarwal and Ariel Herbert-Voss and Gretchen Krueger and Tom Henighan and Rewon Child and Aditya Ramesh and Daniel M. Ziegler and Jeffrey Wu and Clemens Winter and Christopher Hesse and Mark Chen and Eric Sigler and Mateusz Litwin and Scott Gray and Benjamin Chess and Jack Clark and Christopher Berner and Sam McCandlish and Alec Radford and Ilya Sutskever and Dario Amodei},
  title     = {Language Models Are Few-Shot Learners},
  booktitle = {Proc. Adv. Neural Inf. Process. Syst. (NeurIPS)},
  year      = {2020}
}

@article{chen2021codex,
  author  = {Mark Chen and Jerry Tworek and Heewoo Jun and Qiming Yuan and Henrique Ponde de Oliveira Pinto and Jared Kaplan and others},
  title   = {Evaluating Large Language Models Trained on Code},
  journal = {arXiv preprint arXiv:2107.03374},
  year    = {2021}
}

@inproceedings{wei2022cot,
  author    = {Jason Wei and Xuezhi Wang and Dale Schuurmans and Maarten Bosma and Brian Ichter and Fei Xia and Ed H. Chi and Quoc V. Le and Denny Zhou},
  title     = {Chain-of-Thought Prompting Elicits Reasoning in Large Language Models},
  booktitle = {Proc. Adv. Neural Inf. Process. Syst. (NeurIPS)},
  year      = {2022}
}

@article{shinn2023reflexion,
  author  = {Noah Shinn and Federico Cassano and Edward Berman and Ashwin Gopinath and Karthik Narasimhan and Shunyu Yao},
  title   = {Reflexion: Language Agents with Verbal Reinforcement Learning},
  journal = {arXiv preprint arXiv:2303.11366},
  year    = {2023}
}

@inproceedings{yao2023react,
  author    = {Shunyu Yao and Jeffrey Zhao and Dian Yu and Nan Du and Izhak Shafran and Karthik Narasimhan and Yuan Cao},
  title     = {{ReAct}: Synergizing Reasoning and Acting in Language Models},
  booktitle = {Proc. Int. Conf. Learn. Representations (ICLR)},
  year      = {2023}
}

@article{wang2022selfconsistency,
  author  = {Xuezhi Wang and Jason Wei and Dale Schuurmans and Quoc V. Le and Ed H. Chi and Sharan Narang and Aakanksha Chowdhery and Denny Zhou},
  title   = {Self-Consistency Improves Chain-of-Thought Reasoning in Language Models},
  journal = {arXiv preprint arXiv:2203.11171},
  year    = {2022}
}

@article{liu2023agentbench,
  author  = {Xiao Liu and Hao Yu and Hanchen Zhang and Yifan Xu and Xuanyu Lei and Hanyu Lai and Yu Gu and Hangliang Ding and Kaiwen Men and Kejuan Yang and Shudan Zhang and Xiang Deng and Aohan Zeng and Zhengxiao Du and Chenhui Zhang and Sheng Shen and Tianjun Zhang and Yu Su and Huan Sun and Minlie Huang and Yuxiao Dong and Jie Tang},
  title   = {{AgentBench}: Evaluating {LLMs} as Agents},
  journal = {arXiv preprint arXiv:2308.03688},
  year    = {2023}
}

@misc{yang2023autogpt,
  author       = {Toran Bruce Richards and others},
  title        = {{Auto-GPT}: An Autonomous {GPT-4} Experiment},
  howpublished = {\url{https://github.com/Significant-Gravitas/Auto-GPT}},
  year         = {2023}
}

@inproceedings{lewis2020rag,
  author    = {Patrick Lewis and Ethan Perez and Aleksandra Piktus and Fabio Petroni and Vladimir Karpukhin and Naman Goyal and Heinrich K{\"u}ttler and Mike Lewis and Wen-tau Yih and Tim Rockt{\"a}schel and Sebastian Riedel and Douwe Kiela},
  title     = {Retrieval-Augmented Generation for Knowledge-Intensive {NLP} Tasks},
  booktitle = {Proc. Adv. Neural Inf. Process. Syst. (NeurIPS)},
  year      = {2020}
}

@inproceedings{hendrycks2021apps,
  author    = {Dan Hendrycks and Steven Basart and Saurav Kadavath and Mantas Mazeika and Akul Arora and Ethan Guo and Collin Burns and Samir Puranik and Horace He and Dawn Song and Jacob Steinhardt},
  title     = {Measuring Coding Challenge Competence with {APPS}},
  booktitle = {Proc. Int. Conf. Learn. Representations (ICLR)},
  year      = {2021}
}

@inproceedings{feng2020codebert,
  author    = {Zhangyin Feng and Daya Guo and Duyu Tang and Nan Duan and Xiaocheng Feng and Ming Gong and Linjun Shou and Bing Qin and Ting Liu and Daxin Jiang and Ming Zhou},
  title     = {{CodeBERT}: A Pre-Trained Model for Programming and Natural Languages},
  booktitle = {Findings of EMNLP},
  year      = {2020}
}

@article{wang2021codet5,
  author  = {Yue Wang and Weishi Wang and Shafiq Joty and Steven C. H. Hoi},
  title   = {{CodeT5}: Identifier-aware Unified Pre-trained Encoder-Decoder Models for Code Understanding and Generation},
  journal = {arXiv preprint arXiv:2109.00859},
  year    = {2021}
}

@article{schick2023toolformer,
  author  = {Timo Schick and Jane Dwivedi-Yu and Roberto Dess{\`i} and Roberta Raileanu and Maria Lomeli and Luke Zettlemoyer and Nicola Cancedda and Thomas Scialom},
  title   = {Toolformer: Language Models Can Teach Themselves to Use Tools},
  journal = {arXiv preprint arXiv:2302.04761},
  year    = {2023}
}

@article{gao2023pal,
  author  = {Luyu Gao and Aman Madaan and Shuyan Zhou and Uri Alon and Pengfei Liu and Yiming Yang and Jamie Callan and Graham Neubig},
  title   = {{PAL}: Program-Aided Language Models},
  journal = {arXiv preprint arXiv:2211.10435},
  year    = {2022}
}

@misc{chen2018executionguided,
  author       = {Xinyun Chen and Chang Liu and Dawn Song},
  title        = {Execution-Guided Neural Program Synthesis},
  howpublished = {OpenReview},
  year        = {2018}
}

@article{li2022alphacode,
  author  = {Yujia Li and David Choi and Junyoung Chung and Nate Kushman and Julian Schrittwieser and R{\'e}mi Leblond and Tom Eccles and James Keeling and Felix Gimeno and Agustin Dal Lago and Thomas Hubert and Peter Choy and Cyprien de Masson d'Autume and Igor Babuschkin and Xinyun Chen and Po-Sen Huang and Johannes Welbl and Sven Gowal and Alexey Cherepanov and James Molloy and Daniel J. Mankowitz and Esme Sutherland Robson and Pushmeet Kohli and Nando de Freitas and Koray Kavukcuoglu and Oriol Vinyals},
  title   = {Competition-Level Code Generation with {AlphaCode}},
  journal = {Science},
  year    = {2022}
}

@article{chowdhery2022palm,
  author  = {Aakanksha Chowdhery and others},
  title   = {{PaLM}: Scaling Language Modeling with Pathways},
  journal = {arXiv preprint arXiv:2204.02311},
  year    = {2022}
}

@article{nijkamp2022codegen,
  author  = {Erik Nijkamp and Bo Pang and Hiroaki Hayashi and Lifu Tu and Huan Wang and Yingbo Zhou and Silvio Savarese and Caiming Xiong},
  title   = {{CodeGen}: An Open Large Language Model for Code with Multi-Turn Program Synthesis},
  journal = {arXiv preprint arXiv:2203.13474},
  year    = {2022}
}

@article{fried2022incoder,
  author  = {Daniel Fried and Armen Aghajanyan and Jessy Lin and Sida Wang and Eric Wallace and Freda Shi and Ruiqi Zhong and Wen-tau Yih and Luke Zettlemoyer and Mike Lewis},
  title   = {{InCoder}: A Generative Model for Code Infilling and Synthesis},
  journal = {arXiv preprint arXiv:2204.05999},
  year    = {2022}
}

@article{austin2021mbpp,
  author  = {Jacob Austin and Augustus Odena and Maxwell Nye and Maarten Bosma and Henryk Michalewski and David Dohan and Ellen Jiang and Carrie Cai and Michael Terry and Quoc Le and Charles Sutton},
  title   = {Program Synthesis with Large Language Models},
  journal = {arXiv preprint arXiv:2108.07732},
  year    = {2021}
}

@article{chen2023selfdebug,
  author  = {Xinyun Chen and Maxwell Lin and Nathanael Sch{\"a}rli and Denny Zhou},
  title   = {Teaching Large Language Models to Self-Debug},
  journal = {arXiv preprint arXiv:2304.05128},
  year    = {2023}
}

@article{huang2023agentcoder,
  author  = {Dong Huang and Jie M. Zhang and Michael Luck and Qingwen Bu and Yuhao Qing and Heming Cui},
  title   = {{AgentCoder}: Multi-Agent-Based Code Generation with Iterative Testing and Optimisation},
  journal = {arXiv preprint arXiv:2312.13010},
  year    = {2023}
}

@article{islam2024mapcoder,
  author  = {Md Ashraful Islam and Mohammed Eunus Ali and Md Rizwan Parvez},
  title   = {{MapCoder}: Multi-Agent Code Generation for Competitive Problem Solving},
  journal = {arXiv preprint arXiv:2405.11403},
  year    = {2024}
}

@article{hong2023metagpt,
  author  = {Sirui Hong and Mingchen Zhuge and Jonathan Chen and Xiawu Zheng and Yuheng Cheng and Ceyao Zhang and Jinlin Wang and Zili Wang and Steven Ka Shing Yau and Zijuan Lin and Liyang Zhou and Chenyu Ran and Lingfeng Xiao and Chenglin Wu and J{\"u}rgen Schmidhuber},
  title   = {{MetaGPT}: Meta Programming for a Multi-Agent Collaborative Framework},
  journal = {arXiv preprint arXiv:2308.00352},
  year    = {2023}
}

@article{wu2023autogen,
  author  = {Qingyun Wu and Gagan Bansal and Jieyu Zhang and Yiran Wu and Beibin Li and Erkang Zhu and Li Jiang and Xiaoyun Zhang and Shaokun Zhang and Jiale Liu and Ahmed Hassan Awadallah and Ryen W. White and Doug Burger and Chi Wang},
  title   = {{AutoGen}: Enabling Next-Gen LLM Applications via Multi-Agent Conversation},
  journal = {arXiv preprint arXiv:2308.08155},
  year    = {2023}
}

@article{deepseekv2coder,
  author  = {{DeepSeek-AI}},
  title   = {{DeepSeek-Coder-V2}: Breaking the Barrier of Closed-Source Models in Code Intelligence},
  journal = {arXiv preprint arXiv:2406.11931},
  year    = {2024}
}

@article{hui2024qwen25coder,
  author  = {Binyuan Hui and Jian Yang and Zeyu Cui and Jiaxi Yang and Dayiheng Liu and Lei Zhang and Tianyu Liu and Jiajun Zhang and Bowen Yu and Keming Lu and Kai Dang and Yang Fan and Yichang Zhang and An Yang and Rui Men and Fei Huang and Bo Zheng and Yibo Miao and Shanghaoran Quan and Yunlong Feng and Xingzhang Ren and Xuancheng Ren and Jingren Zhou and Junyang Lin},
  title   = {{Qwen2.5-Coder} Technical Report},
  journal = {arXiv preprint arXiv:2409.12186},
  year    = {2024}
}

@misc{qwen3coder,
  author       = {{Qwen Team}},
  title        = {{Qwen3-Coder}: Agentic Code Generation with Mixture-of-Experts},
  howpublished = {Technical report; weights at \url{https://huggingface.co/Qwen/Qwen3-Coder-30B-A3B-Instruct}},
  year         = {2025}
}

@misc{gptoss,
  author       = {{OpenAI}},
  title        = {{gpt-oss}: An Open-Weight Language Model from {OpenAI}},
  howpublished = {Apache-2.0 weights at \url{https://huggingface.co/openai/gpt-oss-20b}},
  year         = {2025}
}

@misc{judgesense2026,
  title         = {{JudgeSense}: A Benchmark for Prompt Sensitivity in {LLM-as-a-Judge} Systems},
  author        = {Bellibatlu, Rohith Reddy and Raff, Edward and Zhang, Wenbin},
  year          = {2026},
  howpublished  = {arXiv preprint arXiv:2604.23478},
  eprint        = {2604.23478},
  archivePrefix = {arXiv},
  primaryClass  = {cs.CL},
  doi           = {10.48550/arXiv.2604.23478},
  url           = {https://arxiv.org/abs/2604.23478}
}

@misc{langchain,
  author       = {{LangChain Authors}},
  title        = {{LangChain}: Building applications with {LLMs} through composability},
  howpublished = {\url{https://docs.langchain.com/}},
  year         = {2024},
  note         = {Open-source software}
}

@misc{langgraph,
  author       = {{LangGraph Authors}},
  title        = {{LangGraph}: A library for building stateful, multi-actor applications with {LLMs}},
  howpublished = {\url{https://github.com/langchain-ai/langgraph}},
  year         = {2024},
  note         = {Open-source software}
}

@misc{chroma,
  author       = {{Chroma Authors}},
  title        = {{Chroma}: The {AI}-native open-source embedding database},
  howpublished = {\url{https://docs.trychroma.com/}},
  year         = {2024},
  note         = {Open-source software}
}

@misc{ollama,
  author       = {{Ollama Authors}},
  title        = {{Ollama}: Run Large Language Models locally},
  howpublished = {\url{https://ollama.com/}},
  year         = {2024},
  note         = {Open-source software}
}

\end{document}